\documentclass{article}

\usepackage{microtype}
\usepackage{graphicx}
\usepackage{subfigure}
\usepackage{booktabs}

\usepackage{hyperref}
\hypersetup{
  pdftitle={Mixture of Training: Recombining Small-Scale Scaffolded Pretraining Runs into a Larger Language Model},
  pdfauthor={Mohammed Sabry, Sean Augenstein, Keith Rush, and Lucio Dery},
  pdfsubject={Accepted at the Workshop on Methods and Opportunities at Small Scale (MOSS), COLM 2026},
  pdfkeywords={language model training, modular training, small-scale pretraining, compute efficiency, MOSS, COLM}
}

\usepackage[accepted]{MOSS_camera_ready}

\makeatletter
\renewcommand{\ICML@appearing}{\textit{Accepted at the Workshop on Methods and Opportunities at Small Scale (MOSS), COLM 2026}.}
\makeatother

\usepackage{amsmath}
\usepackage{amssymb}
\usepackage{mathtools}
\usepackage{amsthm}

\usepackage[capitalize,noabbrev]{cleveref}

\theoremstyle{plain}

\theoremstyle{definition}

\theoremstyle{remark}

\usepackage[disable,textsize=tiny]{todonotes}

\icmltitlerunning{Mixture of Training}

\begin{document}

\onecolumn

\icmltitle{Mixture of Training: Recombining Small-Scale Scaffolded Pretraining Runs into a Larger Language Model}


\begin{icmlauthorlist}
\icmlauthor{Mohammed Sabry}{google,dcu}
\icmlauthor{Sean Augenstein}{google}
\icmlauthor{Keith Rush}{google}
\icmlauthor{Lucio Dery}{google}
\end{icmlauthorlist}

\icmlaffiliation{google}{Google, Mountain View, CA, USA}
\icmlaffiliation{dcu}{School of Computing, Dublin City University, Dublin, Ireland}

\icmlcorrespondingauthor{Mohammed Sabry}{mhmd.sabry.ab@gmail.com}
\icmlcorrespondingauthor{Sean Augenstein}{saugenst@google.com}

\printAffiliationsAndNotice{This work was conducted while Mohammed Sabry was a PhD Student Researcher at Google.}

\icmlkeywords{language model training, modular training, small-scale pretraining, compute efficiency, MOSS, COLM}

\vskip 0.3in


\begin{abstract}
We ask whether language-model pre-training can be decomposed into smaller, independently trainable jobs that can later be recomposed into a coherent larger model. We introduce \emph{Mixture of Training} (MoT), a scaffolded modular pre-training procedure that partitions a target Transformer into contiguous layer blocks, trains each block inside a frozen pretrained aligner scaffold, and then recomposes the trained blocks with an optional short end-to-end adaptation pass. On a 1.3B-parameter Gemma-style model trained on \textsc{C4}, MoT provides a small-scale proof of mechanism: independently trained depth slices can be recomposed into a usable language model, and a quality-parity schedule reaches the same reported perplexity as the monolithic baseline. This parity setting processes more aggregate tokens and has a shorter idealized layer-equivalent critical path after aligner preparation; its effective compute advantage depends on reusing the aligner across runs. We therefore present MoT not as a general replacement for monolithic pre-training, but as a small-scale framework for studying whether scaffolded sub-runs can act as reusable training units.
\end{abstract}

\section{Introduction}

Large language model training is usually organized as one monolithic end-to-end optimization run. This creates a scaling bottleneck: all layers must be trained together, failures affect the whole run, and every improvement requires restarting or continuing a large coupled system. We ask whether pre-training can instead be decomposed into smaller, independently trainable jobs that can later be recomposed into a coherent larger model. This question is especially natural in the small-scale regime targeted by MOSS: if small runs can serve as reusable scientific units, they may make training research cheaper, more reproducible, and easier to iterate on.

We introduce \emph{Mixture of Training} (MoT), a scaffolded modular pre-training method. MoT partitions a target Transformer into contiguous layer blocks, trains each block inside a frozen pretrained \emph{aligner} scaffold, and then recomposes the trained blocks into the full target model. The aligner supplies a stable representational interface during independent block training, so the blocks can be optimized without exchanging gradients while still remaining compatible at composition time.

Our study asks whether this decomposition can preserve language-model quality while changing the compute, token-exposure, and critical-path trade-off. On a 12-layer, 1.3B-parameter Gemma-style language model trained on \textsc{C4}, MoT provides a small-scale proof of mechanism: independently trained depth slices can be recomposed into a coherent model, and a quality-parity schedule reaches the same reported perplexity as the monolithic baseline after end-to-end adaptation. This setting processes more aggregate tokens and has a shorter idealized layer-equivalent critical path after aligner preparation, but its effective compute advantage depends on amortizing the pretrained aligner across reuse cases. A lower-compute MoT schedule remains below the monolithic EFLOP budget even when the aligner is fully charged, but reaches slightly worse perplexity. Ablations show that the aligner is essential; under the tested aligner setting, disjoint submodel data streams improve cold composition; and increasing the number of splits trades quality for additional efficiency.

Our contributions are:
(i) a scaffolded modular training procedure that trains target depth slices independently while preserving a shared representational interface;
(ii) a small-scale proof-of-mechanism study showing that independently trained slices can be recomposed into a 1.3B-parameter language model;
(iii) compute, token-exposure, and estimated critical-path accounting for cold composition, adapted composition, and quality-parity schedules, including explicit treatment of aligner amortization; and
(iv) ablations showing how the aligner, data partitioning, and split count shape the observed quality--efficiency trade-off in this setting.

\paragraph{Positioning.}
MoT most directly adapts Deep Incubation's scaffolded module-training pattern~\citep{ni2023deepincubationtraininglarge} to autoregressive language-model pre-training, examining flexible aligner depth, separately sampled submodel streams, end-to-end adaptation, and compute amortization. Unlike progressive growth, its target blocks train in parallel; unlike model averaging, distinct depth slices form a larger model; and unlike post-hoc stitching, compatibility is encouraged during training. DiffusionBlocks~\citep{shing2026diffusionblocksblockwiseneuralnetwork} also trains blocks independently but uses a diffusion-denoising interpretation, whereas MoT retains next-token prediction and an aligner scaffold. Appendix~\ref{app:related} provides broader comparisons.

\section{Mixture of Training}
\label{sec:mot_pretraining}

Let the target Transformer be written as a composition of \(K\) contiguous layer blocks,
\[
F = f_K \circ f_{K-1} \circ \cdots \circ f_1 .
\]
Each \(f_i\) is a target \emph{submodel}. MoT trains these submodels independently but not in isolation. It introduces a pretrained aligner \(A = a_K \circ \cdots \circ a_1\), sliced into the same number of blocks and shape-compatible with the target.

For each target block \(f_i\), MoT constructs a scaffolded network
\[
S_i =
a_K \circ \cdots \circ a_{i+1} \circ f_i \circ a_{i-1} \circ \cdots \circ a_1 ,
\]
where only \(f_i\) is trainable and all aligner slices are frozen. Each scaffold is optimized with the standard next-token prediction loss. Thus, every target block learns inside the representational context supplied by the same aligner, but no gradients are exchanged between target blocks during scaffolded training.

\paragraph{Interface constraints.}
The aligner is required to be shape-compatible with the target model: it uses the same global width, attention-head dimensionality, feed-forward width, token embedding space, and output head. This lets aligner slices and target slices be swapped without adding projection or stitching layers. In this first study we restrict \(f_i\) to equal or near-equal contiguous layer blocks. Non-contiguous, neuron-level, or expert-level splits are possible, but they introduce additional permutation or projection problems and are left to future work.

\paragraph{Embedding and output handling.}
Each scaffold reuses the aligner's token embeddings and output head; only \(f_i\) is updated. The recomposed model retains these shared components but replaces the aligner's Transformer slices with the trained target blocks, and all reported perplexities are measured on this recomposed model.

MoT has three stages. \textbf{Stage 0} prepares or selects the aligner and partitions both aligner and target into corresponding blocks. \textbf{Stage 1} trains all scaffolded networks \(S_i\) in parallel, updating only the target block in each scaffold. \textbf{Stage 2} discards the aligner, recomposes \(\hat{F}=f_K\circ\cdots\circ f_1\), and optionally applies a short end-to-end adaptation pass. We call the recomposed model before Stage 2 adaptation the \emph{cold-composed} model; its quality directly measures how well scaffolded training aligned the independently trained blocks.

\begin{figure}[t]
    \centering
    \includegraphics[width=0.9\linewidth]{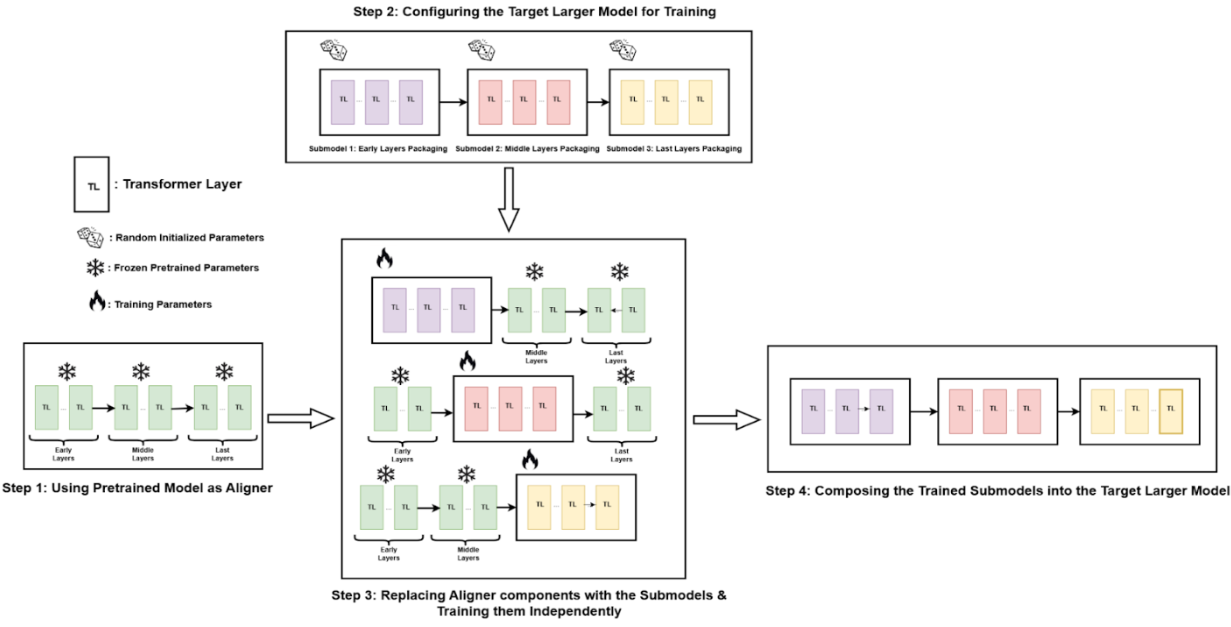}
    \caption{Mixture of Training (MoT). The illustrated preparation and slicing steps form Stage~0; scaffolded block training is Stage~1; and recomposition with optional end-to-end adaptation is Stage~2. Only the target block in each scaffold is updated.}
    \label{fig:mot_approach}
\end{figure}

\section{Experiments and Results}
\label{sec:exp_and_results}

We evaluate MoT on the English portion of \textsc{C4}. The target model is a 12-layer, 1.3B-parameter decoder-only Transformer following the Gemma-1-2B width configuration: 256k token vocabulary, 2048-dimensional embeddings, RoPE, multi-query attention, and 16384-dimensional feed-forward layers. Optimization uses AdamW with the same batch size across the baseline and MoT runs, but Stage~1 uses a different post-warm-up learning-rate schedule; full details appear in Appendix~\ref{app:aligner}. The monolithic baseline is trained end-to-end for 128k updates, processing 33.6B tokens and reaching perplexity 15.0 at 268.4 EFLOPs. Unless otherwise noted, MoT uses \(K=2\) target blocks, a 4-layer aligner, disjoint data streams for the two submodels, and evaluation is always performed on the recomposed target model rather than on individual scaffolds.

We report perplexity, the exponential of mean token-level cross-entropy on held-out \textsc{C4}. Baseline and MoT runs share the data source, optimizer family, and batch setting, with stage-specific learning-rate schedules reported in Appendix~\ref{app:aligner}. Because Stage~1 reorganizes compute and exposure across parallel submodels, we report training and fully charged EFLOPs, aggregate tokens, and an idealized layer-equivalent critical-path estimate rather than presenting the latter as measured wall-clock or equal-hardware speedup.

Table~\ref{tab:main_results} summarizes the main quality--compute trade-off across three MoT schedules. \emph{Cold composition} trains the two 6-layer submodels for 50k updates each and evaluates the recomposed model before any end-to-end adaptation. \emph{MoT + 15k adaptation} adds a short Stage-2 pass to measure how much mismatch remains after independent scaffolded training. \emph{MoT quality parity} reinvests part of the saved compute by extending submodel training to 75k updates and using a 30k adaptation pass. Because the 4-layer aligner costs 29.7 EFLOPs once, we report both direct MoT training EFLOPs and fully charged EFLOPs that assign the entire aligner cost to one run. We describe amortized reuse separately below rather than choosing a particular reuse count in the table. The critical-path derivation appears in Appendix~\ref{app:wallclock}.

\begin{table}[h]
\centering
\caption{Main results. EFLOPs follow Appendix~\ref{app:compute}; ``Train EF'' excludes aligner preparation, while ``Full-charge EF'' includes its entire 29.7 EFLOP cost. Tokens are aggregated across the schedule. The idealized layer-equivalent critical path assumes concurrent Stage~1 jobs after aligner preparation and is not a measured equal-hardware speedup.}
\label{tab:main_results}
\begin{tabular}{lccccc}
\toprule
Model / schedule & PPL $\downarrow$ & Train EF & Full-charge EF & Tokens (B) & Critical-path estimate \\
\midrule
Monolithic baseline & 15.0 & 268.4 & 268.4 & 33.6 & 1.0$\times$ \\
MoT cold composition & 19.3 & 128.2 & 157.9 & 26.2 & 4.2$\times$ \\
MoT + 15k adaptation & 15.9 & 159.7 & 189.4 & 30.1 & 2.8$\times$ \\
MoT quality parity & 15.0 & 255.3 & 285.0 & 47.1 & 1.7$\times$ \\
\bottomrule
\end{tabular}
\end{table}

If the aligner is charged fully to a single run, the quality-parity schedule costs \(255.3 + 29.7 = 285.0\) EFLOPs, above the 268.4 EFLOP monolithic baseline. Quality parity is therefore not compute-saving in the fully charged single-run setting. More generally, if the same 29.7 EFLOP aligner is reused across \(R\) independent, shape-compatible target-model training runs, the effective quality-parity cost per run is
\[
255.3 + \frac{29.7}{R}.
\]
Here, \(R\) counts independent target-model runs that reuse the same aligner, not checkpoints or ablations from one training effort. The effective cost falls below the monolithic baseline for \(R \geq 3\), so we interpret the quality-parity result as evidence for an amortized-reuse regime rather than an unconditional efficiency gain. The \(50\text{k}+15\text{k}\) schedule remains below the full 128k-step baseline budget even when the aligner is fully charged, reaching PPL 15.9 at \(128.2 + 31.5 + 29.7 = 189.4\) EFLOPs. However, because we do not report a monolithic checkpoint trained or tuned at the same 189.4 EFLOP budget, this comparison does not establish an equal-compute advantage. Finally, the critical-path estimates assume that the two Stage~1 scaffold jobs run concurrently and exclude aligner preparation; they characterize time-to-result under additional parallel hardware, not measured wall-clock time under equal resources.

The aligner is critical in our setting (Table~\ref{tab:main_ablation}; full matrix in Appendix~\ref{app:ablate}). Without it, cold-composition quality drops sharply. Disjoint streams improve PPL under the tested 4-layer aligner but worsen it without an aligner. Increasing \(K\) from 2 to 4 lowers compute but worsens cold-composition PPL, exposing a quality--efficiency trade-off.

\begin{table}[h]
\centering
\caption{Compact cold-composition ablations. As in the ``Full-charge EF'' column of Table~\ref{tab:main_results}, this table charges the full 29.7 EFLOP aligner cost where present.}
\label{tab:main_ablation}
\begin{tabular}{lccc}
\toprule
Setting & PPL $\downarrow$ & EFLOPs & Takeaway \\
\midrule
No aligner, \(K=2\), shared data & 38.9 & 104.8 & incompatible \\
Aligner, \(K=2\), shared data & 20.3 & 158 & scaffold helps \\
Aligner, \(K=2\), disjoint data & 19.3 & 158 & best cold PPL \\
Aligner, \(K=4\), disjoint data & 24.8 & 110 & cheaper, worse PPL \\
\bottomrule
\end{tabular}
\end{table}

These ablations provide behavioural evidence for the role of the aligner, but they do not by themselves fully diagnose the internal mechanism. The large degradation without an aligner is consistent with an interface-mismatch failure: independently trained depth slices do not necessarily produce hidden states that the next slice can use. The frozen aligner plausibly reduces this mismatch by making every slice train against the same surrounding representational context. Under the tested aligner settings, disjoint data streams improve the cold-composed model, suggesting that the recomposed target can benefit from broader token exposure when the scaffold preserves a shared interface. Increasing \(K\) gives more parallelism and lower Stage-1 compute, but each trainable block becomes smaller and the number of interfaces increases, so cold-composition quality drops. A fuller mechanistic account would require direct interface diagnostics, such as hidden-state similarity across boundaries, activation-norm drift, CKA/SVCCA comparisons, or layer-wise loss probes before and after recomposition.

\section{Discussion and Limitations}
\label{sec:discussion}

The experiments support three conclusions. First, independently trained submodels can be recomposed into a coherent language model when trained inside a shared aligner scaffold; without the scaffold, cold composition degrades sharply. Second, the remaining mismatch is mostly recoverable rather than catastrophic: a short adaptation pass closes most of the gap, and a longer quality-parity schedule reaches the same reported perplexity as the monolithic baseline. Third, MoT changes the engineering shape of pre-training. Instead of one coupled run, it creates smaller jobs that can be scheduled, restarted, and ablated independently, and potentially reused across compatible training efforts, making it especially suitable for small-scale training research.

The current study is intentionally small-scale. It uses one model family, one dataset, and a limited set of schedules, and reports perplexity rather than downstream reasoning, factuality, calibration, or robustness benchmarks. We also do not yet include all compute-matched monolithic controls, such as baselines matched for fully charged MoT EFLOPs, aggregate token exposure, or estimated critical-path budget, nor do we report measured wall-clock time under equal hardware resources. Consequently, the current evidence demonstrates modular composability in this setting and identifies an amortized-reuse scheduling regime, rather than a uniform compute or equal-hardware speed advantage. As a proof of mechanism, MoT opens a concrete design space for reusable scaffolded sub-runs and their quality--compute--scheduling trade-offs.

\bibliography{MoT}
\bibliographystyle{icml2025}

\appendix

\section{Additional Related Work}
\label{app:related}
\paragraph{Progressive / Growth Strategies:}
Progressive depth or capacity growth \citep{chen-etal-2022-bert2bert, Yao2023MaskedSG, Du2024StackingYT} trains a smaller network first and incrementally adds layers or width, continuing end-to-end optimization of the enlarged model; this saves early compute but is inherently sequential and any early representational biases propagate forward. In contrast, MoT trains all planned layer groups in parallel from the outset under a shared aligner interface, turning depth segmentation into a parallel rather than temporal schedule and decoupling failures or slow convergence in one segment from others.

\paragraph{Model Merging and Weight Averaging:}
Model soups \citep{wortsman2022modelsoupsaveragingweights} (simple parameter averaging of multiple finetuned checkpoints) and earlier weight-averaging / merging techniques combine identically-shaped models to improve accuracy or robustness without inference cost increase, relying on the empirical flatness / basin connectivity of fine-tuned solutions. However, merging does not enlarge capacity and can lose complementary information if models are not well aligned; alignment is incidental, arising from starting from a common pretrained seed. In MoT each submodel occupies a distinct layer subset so assembling increases overall depth coverage (capacity) beyond any single trained component, and alignment can be actively enforced throughout submodel training by the frozen aligner interface, rather than assumed at the moment of averaging.

\paragraph{Parallel Learning and Loosely-Coupled Training:}

Weight averaging and model merging techniques have also found homes at training time in Local SGD (dating to at least ~\citep{NIPS2009_d81f9c1b}) and its many related methods. These methods share the core technique of independently evolving models before leveraging some kind of parameter- or model-delta-space merging. In the most extreme cases this takes the form of a one-shot average, in the style of model souping; often the technique is applied iteratively. MoT bears some resemblance to these local-training based methods, but crucially differs in the manner of composition, with the explicit goal of the composed model having higher capacity than any of its locally-trained slices.

\paragraph{Model Stitching and Representation Alignment:}
Model stitching research \citep{DBLP:journals/corr/abs-2106-07682, hernandez2023modelstitchinglookingfunctional} examines substituting intermediate layers between pretrained networks and inserting ``stitch'' projection layers to reconcile activation spaces, enabling analysis or hybrid reuse; success depends on latent space compatibility and often requires additional bridging parameters. MoT avoids explicit stitching layers at assembly time because its aligner enforces a stable intermediate representation contract during every submodel's training, yielding direct composability without post-hoc adapters, and allowing frozen aligner parts to be discarded after composition.

\paragraph{Divide-and-Conquer and Block-wise Training:}
Deep Incubation \citep{ni2023deepincubationtraininglarge, Lo2024m2mKDMK} divides large vision transformers into submodules trained to replace segments of a small global ``meta model,'' improving ImageNet accuracy and reducing training cost; the meta model implicitly links modules. MoT applies the same broad divide-and-conquer pattern to autoregressive language-model pre-training. The present study emphasizes five axes in this setting: (i) a \emph{flexible} aligner scale (not locked to one layer per module);
(ii) pretrained aligners as scaffold interfaces, with instruction- or value-aligned variants left as future work;
(iii) multiple split granularities and separately sampled data streams per submodel;
(iv) explicit FLOP, memory, critical-path, and amortization accounting; and
(v) end-to-end adaptation after cold composition.

Recent methods such as DiffusionBlocks~\citep{shing2026diffusionblocksblockwiseneuralnetwork} also train blocks independently, but cast residual blocks as diffusion denoising steps over assigned noise ranges; MoT instead keeps the autoregressive next-token objective and uses an aligner scaffold to make independently trained depth slices composable.

\paragraph{Hardware Fault Tolerance and Training Failures:}
Large monolithic training runs face hardware faults, instability, and silent performance regressions; recent analyses \citep{jiang2025l4diagnosinglargescalellm} of large-scale logs underscore the operational complexity and failure modes in multi-week LLM training. By decomposing the workload into independently restartable submodel jobs, MoT could reduce the amount of work affected by a failure and simplify recovery because only the affected scaffold job would need to resume. We do not measure fault-recovery behavior here, so this remains a prospective systems benefit rather than an empirical result.

\paragraph{Summary:}
Existing approaches ensure compatibility of distributed components through continuous activation and gradient synchronization (data/model/pipeline parallelism), post-hoc adapters (stitching), parameter merging (model soups and local SGD), or a shared meta-model scaffold (Deep Incubation). Relative to this literature, MoT studies the scaffolded divide-and-conquer pattern for autoregressive language-model pre-training, including flexible aligner depth, disjoint submodel data streams, end-to-end adaptation, and explicit FLOP, memory, critical-path, and amortization accounting. Behaviour-specific aligners and reuse across independent target-model runs remain directions for future evaluation.

\section{Compute Cost Analysis}
\label{app:compute}

\subsection{Baseline pre-training}

For a transformer with \(N\) non-embedding parameters and a training
corpus of \(D\) tokens, the approximate forward--backward cost is
\[
  C_{\text{baseline}}
  \;\approx\;
  6\,N\,D ,
\]
where the factor six is the standard training-FLOP approximation: roughly
\(2ND\) FLOPs for the forward pass and \(4ND\) FLOPs for the backward pass. This layer-dominated approximation excludes embedding lookup, output projection/softmax, input-pipeline, and other fixed costs; the 256k-vocabulary output head may therefore make the reported values optimistic as full-system compute estimates.
When the token--to--parameter ratio follows the Chinchilla rule \citep{hoffmann2022trainingcomputeoptimallargelanguage}
\(M = D / N = 20\), the expression reduces to
\[
  C_{\text{baseline}} \approx 120\,N^{2}.
\]

\subsection{MoT pre-training}

MoT partitions the target model into \(K\) trainable submodels of size
\(N_{\!M}\) and introduces an aligner of size \(N_{\!A}\), sliced into the same
\(K\) segments and kept frozen.
Let \(D_{\!M}\) be the token budget per submodel and
\(D_{\text{adapt}}\) the token budget for Stage 2 adaptation.
Using the same layer-dominated approximation, a conservative per-submodel upper bound for Stage~1 is
\[
\begin{aligned}
C_{\text{mot\,submodel}}
 \;&=\;
 2\bigl(N_{\!M} + (K-1)N_{\!A}/K\bigr)D_{\!M}       &&\text{(forward pass)} \\
 &\quad
 +\;4N_{\!M}D_{\!M}                                 &&\text{(weight \& activation gradients)} \\
 &\quad
 +\;2(K-1)N_{\!A}/K\,D_{\!M}                        &&\text{(activation gradients for frozen slices)} .
\end{aligned}
\]
The final activation-gradient term charges every frozen slice as though it lies downstream of the trainable block. In an implementation that stops gradients through upstream frozen slices, this overcounts those upstream slices; we retain it as a transparent conservative bound. Because all \(K\) scaffolded models are trained in parallel, the aggregate Stage 1 cost under this accounting is
\(K \times C_{\text{mot\,submodel}}\).

Stage 2 adapts the recomposed model and costs
\(
  C_{\text{adapt}} \approx 6\,N\,D_{\text{adapt}}.
\)

Hence the overall MoT budget is
\[
  C_{\text{MoT}}
  \;=\;
  C_{\text{align}}
  \;+\;
  K\,C_{\text{mot\,submodel}}
  \;+\;
  C_{\text{adapt}},
\]
where \(C_{\text{align}}\) is the Stage 0 cost.
In many scenarios the aligner can be reused from an off-the-shelf checkpoint,
so we treat \(C_{\text{align}}\) as amortizable. The main text reports direct and fully charged costs, then gives the amortized per-run cost explicitly as a function of the number of independent target-model runs that reuse the aligner.

\paragraph{Numerical example:}
For the 1.3B Gemma-style setting with \(K = 2\) splits and a 4-layer aligner the
reported layer-equivalent cost is:

\begin{itemize}
    \item Stage 0 (aligner): \(\;29.7\) EFLOPs
    \item Stage 1 (submodels): \(\;128.2\) EFLOPs
    \item Stage 2 (15\,k steps): \(\;31.5\) EFLOPs
\end{itemize}

\noindent
The total of \(189.4\) EFLOPs is 29\% below the full 128k-step baseline budget of
268.4 EFLOPs while reaching PPL 15.9, closing most of the cold-composition gap after a 15k-step adaptation. This is not an equal-compute comparison: we do not report a monolithic checkpoint trained or tuned at the same 189.4 EFLOP budget.

\section{Memory Footprint Analysis}
\label{app:memory}

This appendix complements the FLOP and critical-path analysis by accounting for the memory terms introduced by scaffolded training. During Stage~1, each scaffold stores optimizer states and parameter gradients only for the trainable target block, while frozen aligner slices still contribute parameter storage and forward activations. As in Appendix~\ref{app:compute}, \(N\) counts non-embedding Transformer parameters: shared token-embedding and output-head storage must be added for a full implementation-level estimate and may be material with a 256k-token vocabulary. The resulting peak memory therefore depends on the balance between reduced trainable-state storage and additional scaffold activations, as well as implementation choices such as activation checkpointing, recomputation, sharding, and attention kernels.

\subsection{Baseline memory usage}

With FP32 parameters, AdamW optimization and no activation recomputation, a
full end-to-end transformer pretrain requires
\[
\underbrace{4N}_{\text{model}}
+\underbrace{8N}_{\text{AdamW moments}^{\dagger}}
+\underbrace{4N}_{\text{gradients}}
+\underbrace{20\,B\,T\,H\,L}_{\text{activations}}
\quad\text{bytes}.
\]
Here \(N\) is the number of non-embedding parameters,
\(B\) the local batch represented by the footprint,
\(T\) the sequence length,
\(H\) the hidden size and
\(L\) the number of layers.
\par\smallskip
\noindent\(\dagger\) In pure FP32 training, AdamW updates the model parameter directly and stores two additional FP32 moment tensors ($m,v$), for $2\times4=8$ bytes per parameter. A separate FP32 master-weight copy, sometimes used in mixed-precision training, is not assumed here.

\paragraph{Illustrative numerical example:}
The following calculation uses an 18-layer Gemma-2B-scale configuration to illustrate the memory terms; it is not the exact 12-layer, 1.3B headline run or a report of its hardware allocation. For an unsharded logical replica with \(N = 2\text{B}\), \(B = 256\), \(T = 1024\), \(H = 2048\), and \(L = 18\), the baseline footprint is

\[
\text{model}=  8\text{\,GB},\quad
\text{AdamW moments}= 16\text{\,GB},\quad
\text{grads}=  8\text{\,GB},\quad
\text{acts}= 193\text{\,GB},
\]
for a total of roughly \textbf{225~GB} before parameter, optimizer-state, activation, or batch sharding.

\subsection{MoT memory usage}

During Stage 1 each scaffolded model
``frozen aligner slices $+$ trainable submodel''
holds

\[
\bigl(4N_{\!M} + 4(K-1)N_{\!A}/K\bigr)
\;+\;
8N_{\!M} + 4N_{\!M}
\;+\;
20\,B\,T\,H\,L_{\text{hyb}}
\quad\text{bytes},
\]
where \(L_{\text{hyb}}\) denotes the number of layers executed by the scaffold, including the trainable target block and the retained frozen aligner slices. The terms correspond to:

\begin{itemize}
    \item model-parameter storage for the trainable target block and retained frozen aligner slices,
    \item AdamW optimizer states and gradients only for the trainable submodel, since the aligner parameters are frozen, and
    \item forward activations for both the frozen aligner slices and the trainable submodel.
\end{itemize}

\begin{table}[h]
\centering
\caption{Illustrative unsharded memory footprint for an 18-layer Gemma-2B-scale baseline versus one MoT scaffold (\(K=2\)) in Stage~1. Each scaffold trains half of the target layers and retains half of the frozen aligner. This is an accounting example rather than the exact 12-layer, 1.3B headline configuration or a measured per-device footprint.}
\label{tab:memory}
\begin{tabular}{lccccc}
\toprule
 & \textbf{Model} & \textbf{Adam moments} & \textbf{Grads} & \textbf{Acts} & \textbf{Total} \\
\midrule
Baseline (18L) & 8.0\,GB & 16.0\,GB & 8.0\,GB & 193\,GB & 225\,GB \\
MoT, 4-L aligner & 4.8\,GB & 8.0\,GB & 4.0\,GB & 118\,GB & 135\,GB \\
MoT, 8-L aligner & 5.6\,GB & 8.0\,GB & 4.0\,GB & 139\,GB & 157\,GB \\
\bottomrule
\end{tabular}
\end{table}

The table separates the main memory terms rather than reducing them to a single headline claim. MoT reduces optimizer and gradient storage for the trainable component of each scaffold, but the frozen aligner slices add parameter and activation terms. This makes the peak memory footprint implementation-dependent, especially under different activation-checkpointing or recomputation strategies.

With more than two splits, each individual aligner slice is smaller, but each scaffold contains \(K-1\) frozen aligner slices; the total Stage-1 footprint therefore depends on the trade-off between smaller trainable blocks and the retained frozen scaffold depth.

\section{Idealized Layer-Equivalent Critical-Path Analysis} \label{app:wallclock}
The critical-path estimates reported in the main text are idealized schedule-level calculations, not measured end-to-end wall-clock timings. They assume that the \(K\) scaffolded Stage~1 jobs can be scheduled concurrently, so the elapsed time is determined by the slowest scaffold job rather than by the sum of all scaffold jobs. The calculation assigns equal cost to each Transformer layer and omits embedding lookup, output projection/softmax, data loading, communication, and other fixed costs. It is useful for describing the schedule's layer-equivalent dependency path under available parallel hardware, but it should not be read as a same-accelerator-count throughput measurement. A fully measured systems comparison would need to account for hardware allocation, utilization, communication overheads, input-pipeline bottlenecks, activation checkpointing, and implementation-specific kernel efficiency.

A baseline step touches all \(L_{\text{full}}\) layers and costs
\(t_{\text{full}}\) time unit per step. For consistency with the FLOP accounting in Appendix~\ref{app:compute}, we assign one unit to a forward pass, two units to the full backward pass of a trainable layer, and one additional unit to propagating activation gradients through a frozen downstream layer. In the slower of the two \(K=2\) scaffolds, the frozen aligner slice follows the trainable target block and therefore lies on the backward path. The slower-scaffold-to-baseline layer-equivalent step-cost ratio is consequently

\[
c_{\text{scaffold}}
  = \frac{3L_{\text{train}}+2L_{\text{frozen}}}{3L_{\text{full}}} .
\]

In our setup each scaffold in Stage 1 contains \(L_{\text{train}} = 6\)
learnable layers (the split half of the 12-layer target) and
\(L_{\text{frozen}} = 2\) frozen aligner layers
(the 4-layer aligner is split evenly across the two scaffolds). Hence

\[
c_{\text{scaffold}}
  = \frac{3L_{\text{train}} + 2L_{\text{frozen}}}{3L_{\text{full}}}
  = \frac{3\!\times\!6 + 2\!\times\!2}{3\!\times\!12}
  \approx 0.61.
\]

Because the two scaffolds advance in lock-step, Stage 1 lasts \(S_1\)
iterations and Stage 2 lasts \(S_2\).
With a 128 k-step monolithic baseline, the elapsed time is:

\[
T_{\text{MoT}}
  = S_1\,c_{\text{scaffold}}\,t_{\text{full}}
    + S_2\,t_{\text{full}},
\quad
\text{critical-path estimate}
  = \frac{T_{\text{baseline}}}{T_{\text{MoT}}}
  = \frac{128\,\text{k}}
         {S_1\,c_{\text{scaffold}} + S_2}.
\]

Table~\ref{tab:wallclock_sens} plugs in the schedules reported in Table~\ref{tab:main_results}. These are critical-path estimates after aligner preparation, not measurements that include Stage~0 aligner training time.
\begin{table}[h]
\centering
\caption{Idealized layer-equivalent critical-path estimates for the main MoT settings after aligner preparation. The baseline trains for 128k full-model steps. For the 4-layer aligner with \(K=2\), the slower scaffold has \(c_{\text{scaffold}}\approx0.61\). These values assume concurrent Stage~1 jobs and are not measured equal-hardware speedups.}
\label{tab:wallclock_sens}
\begin{tabular}{lccccc}
\toprule
Schedule & $S_1$ & $S_2$ & $c_{\text{scaffold}}$ & Time (k step-equiv.) & Critical-path estimate \\
\midrule
Monolithic baseline & -- & -- & -- & 128.0 & 1.0$\times$ \\
MoT cold composition & 50k & 0k & 0.61 & 30.6 & 4.2$\times$ \\
MoT + 15k adaptation & 50k & 15k & 0.61 & 45.6 & 2.8$\times$ \\
MoT quality parity & 75k & 30k & 0.61 & 75.8 & 1.7$\times$ \\ \bottomrule
\end{tabular}
\end{table}

\section{Aligner Preparation}
\label{app:aligner}

\paragraph{Experimental configuration.}
All runs use a batch size of 256 sequences of length 1024, AdamW with peak learning rate $10^{-3}$, weight decay $10^{-4}$, and $(\beta_1,\beta_2)=(0.9,0.999)$. The baseline, aligner-preparation runs, and Stage~2 adaptation use linear warm-up over the first 10\% of their token budget followed by cosine decay. Stage~1 scaffold training instead uses the same warm-up fraction followed by a constant peak learning rate; a separate three-seed scheduler ablation found similar loss trajectories for warm-up--constant and warm-up--cosine schedules in a smaller 4-layer target setting. The models use post-normalization in the attention and feed-forward blocks and final-logit softening of 30. The main results are individual runs; only the scheduler ablation was replicated.

All aligners share the Gemma-1-2B \citep{gemma1} width configuration but vary in depth and token budget (Table~\ref{tab:aligners}). Except for depth and token budget, aligner-preparation hyperparameters match the baseline.

Because MoT is meant to reuse \emph{off-the-shelf} or behavior-specific guides, we adopt a weak form of data decoupling in this first study: aligners process independently seeded \textsc{C4} dataloader streams containing 10--40B tokens. We did not enforce sample-level de-duplication or measure realized overlap, so these should be understood as separately sampled streams rather than disjoint corpora. This setup tests whether submodels can remain compositionally compatible under an aligner exposed to a partially different sample stream, a minimal proxy for the ``different priors'' scenario we aim to explore in future work.

\begin{table}[h]
\centering
\caption{Aligner configurations and pre-training cost.}
\label{tab:aligners}
\begin{tabular}{lcccc}
\toprule
\textbf{Aligner} & \textbf{\#Params (B)} & \textbf{Token budget} &\textbf{PPL}\,$\downarrow$ & \textbf{FLOPs (EF)} \\
\midrule
4 layers, $M{=}25$ & 0.4 & 10\,B & 20.7 & 29.7 \\
4 layers, $M{=}50$ & 0.4 & 20\,B & 18.4 & 59.4 \\
4 layers, $M{=}100$ & 0.4 & 40\,B & 17.6 & 118.8 \\
6 layers, $M{=}25$ & 0.6 & 15\,B & 17.8& 66.5 \\
8 layers, $M{=}25$ & 0.8 & 20\,B & 16.4& 118.4 \\
\bottomrule
\end{tabular}
\end{table}

\section{Ablation Studies}
\label{app:ablate}

\subsection{Effect of aligner size, split count, and batch sharing}

Table~\ref{tab:full_ablations} summarizes the main ablations.
Models without an aligner or with four splits ($K{=}4$) save more
compute but suffer higher perplexity.
Under the 4-layer aligner, training each submodel on \emph{disjoint}
mini-batches outperforms shared-batch training; without an aligner, the
same change worsens perplexity.

\begin{table}[h]
\centering
\caption{Detailed ablations on aligner, batch sharing and split count for the 12-layer target model. Numbers to the left of ``+'' exclude the one-time \textbf{29.7 EF} aligner cost; the arrow shows the total once that cost is added.}
\label{tab:full_ablations}
\begin{tabular}{llccc}
\toprule
\textbf{Aligner} & \textbf{Batch} & \textbf{Splits (K)} & \textbf{PPL}\,$\downarrow$ & \textbf{Compute (EF)} \\
\midrule
None & Same      & 2 &  38.9 & 104.8 \\
None & Same      & 4 & 478.2 &  48.2 \\
None & Different & 2 &  50.4 & 104.8 \\
None & Different & 4 & 584.0 &  48.2 \\
\midrule
4-layer aligner ($M{=}25$) & Same      & 2 &  20.3 & 128.2 $+\,29.7 \;\rightarrow\; 157.9$ \\
4-layer aligner ($M{=}25$) & Same      & 4 &  28.5 &  80.4 $+\,29.7 \;\rightarrow\; 110.1$ \\
4-layer aligner ($M{=}25$) & Different & 2 &  19.3 & 128.2 $+\,29.7 \;\rightarrow\; 157.9$ \\
4-layer aligner ($M{=}25$) & Different & 4 &  24.8 &  80.4 $+\,29.7 \;\rightarrow\; 110.1$ \\
\bottomrule
\end{tabular}
\end{table}

\paragraph{Observations:}
\begin{itemize}
    \item \textbf{Aligner vs.\ no aligner:}
          Removing the aligner leads to a dramatic quality collapse
          (PPL rising from \(19\!-\!20\) to \(39\!+\)), confirming that the
          aligner's representational scaffold is essential.
    \item \textbf{Shared vs.\ disjoint batches:}
          Under the 4-layer aligner, different mini-batches improve PPL for
          both $K=2$ ($20.3\!\rightarrow\!19.3$) and $K=4$
          ($28.5\!\rightarrow\!24.8$). Without an aligner, they worsen PPL
          ($38.9\!\rightarrow\!50.4$ and $478.2\!\rightarrow\!584.0$). Thus,
          broader token coverage helps in these runs only when the scaffold
          preserves interface compatibility.
    \item \textbf{Split count:}
          Increasing the splits from two to four further reduces Stage 1 compute by approximately 37\% but
          degrades quality unless offset by a stronger adaptation. The trade-off illustrates MoT's
          ability to dial between speed and accuracy.
\end{itemize}

\subsection{MoT Quality Trajectory}\label{app:mot_quality_curve}

\paragraph{Cold-composition quality:}

Table~\ref{tab:stage12} reports C4 validation perplexity and total FLOPs after completing only
\textit{Stage 0} (aligner training) and \textit{Stage 1} (parallel submodel training), \emph{before}
any end-to-end adaptation.
Although no layer sees gradients from the full network, the reassembled model remains within four to five perplexity points of the baseline.
The four-layer $M=50$ setting is one favorable point on the observed frontier:
it reaches PPL 18.9 at 187.6 EFLOPs, approximately 30\% below the fully trained
baseline budget, while the $M=100$ setting reaches the same reported PPL at
higher compute.

\begin{table}[h]
\centering
\caption{Cold Composition performance after Stage\,0\,+\,1 ($K{=}2$ splits, no adaptation). Total FLOPs are shown as \emph{Stage 1 cost + Stage 0 cost $\rightarrow$ summed total}.}
\label{tab:stage12}
\begin{tabular}{lcc}
\toprule
\textbf{Configuration} & \textbf{PPL}\,$\downarrow$ & \textbf{Total FLOPs (EF)} \\
\midrule
Baseline & 15.0   & 268.4 \\
\midrule
MoT, 4-layer aligner ($M{=}25$) & 19.3 & $128.2+\,29.7 \;\rightarrow\; 157.9$ \\
MoT, 4-layer aligner ($M{=}50$) & 18.9 & $128.2 +\,59.4 \;\rightarrow\; 187.6$ \\
MoT, 4-layer aligner ($M{=}100$) & 18.9  & $128.2 +\,118.8 \;\rightarrow\; 247.0$ \\
MoT, 6-layer aligner ($M{=}25$) & 19.3 & $139.8 +\,66.5 \;\rightarrow\; 206.3$ \\
MoT, 8-layer aligner ($M{=}25$) & 19.7 & $151.5 +\,118.4 \;\rightarrow\; 269.9$ \\
\bottomrule
\end{tabular}
\end{table}

Increasing aligner depth or token budget beyond this setting yields no
consistent cold-composition improvement while increasing total compute. These
runs do not identify the mechanism behind this pattern; direct interface
diagnostics would be needed to test whether aligner capacity constrains the
target blocks.

\paragraph{Adaptation benefit:}
Table~\ref{tab:stage123} appends one end-to-end adaptation pass to every
cold-composition run: \textbf{15 k steps}
(\(\sim\!3.9\) B tokens, 31.5 EF) for the 4-layer ($M{=}25,50$) and 6-layer aligners,
and a shorter \textbf{10 k pass}
(\(\sim\!2.6\) B tokens, 21.0 EF) for the largest
4-layer ($M{=}100$) and 8-layer aligners to keep their total compute in
the same ballpark as the baseline.
Even this short adaptation pass closes most of the gap:
with the 4-layer aligner at $M{=}25$ the model reaches
\(\text{PPL}=15.9\) while still saving 29\% of the baseline EFLOPs when the aligner cost is included and 40\% when it is excluded. Additional aligner depth or token budget does not improve adapted PPL in these runs and reduces or reverses the compute savings. Because the largest 4-layer and 8-layer aligners receive only 10k adaptation steps rather than 15k, these rows are not controlled comparisons of aligner capacity alone.

\begin{table}[h]
\centering
\caption{Perplexity after the Stage-2 adaptation pass.
Stage-2 EFLOPs are 31.5 (15 k) or 21.0 (10 k). Total FLOPs are shown as \emph{Stage 2 cost + Stage 1 cost + Stage 0 cost \(\rightarrow\) summed total}.}
\label{tab:stage123}
\begin{tabular}{lccc}
\toprule
\textbf{Configuration} & \textbf{Stage-2 steps} & \textbf{PPL}\,$\downarrow$
 & \textbf{Total EFLOPs} \\
\midrule
Baseline          & ---     & 15.0    & 268.4 \\
\midrule
MoT, 4-layer aligner ($M{=}25$)  & 15 k & 15.9 & $31.5+\,128.2+\,29.7 \;\rightarrow\; 189.4$ \\
MoT, 4-layer aligner ($M{=}50$)  & 15 k & 15.9 & $31.5+\,128.2 +\,59.4 \;\rightarrow\; 219.1$ \\
MoT, 4-layer aligner ($M{=}100$) & 10 k & 16.4 & $21.0+\,128.2 +\,118.8 \;\rightarrow\; 268.0$ \\
MoT, 6-layer aligner ($M{=}25$)  & 15 k & 16.0 & $31.5+\,139.8 +\,66.5 \;\rightarrow\; 237.8$ \\
MoT, 8-layer aligner ($M{=}25$)  & 10 k & 16.7 & $21.0+\,151.5 +\,118.4 \;\rightarrow\; 290.9$ \\
\bottomrule
\end{tabular}
\end{table}

\end{document}